\documentclass[12pt,a4paper, twoside]{article}
\usepackage[utf8]{inputenc}
\usepackage[english]{babel}
\usepackage{libertine}
\usepackage[T1]{fontenc}
\usepackage{amsmath}
\usepackage{amsfonts}
\usepackage{amssymb}
\usepackage{makeidx}
\usepackage{graphicx}
\usepackage{listings}
\usepackage{xcolor}
\usepackage{float}
\usepackage[tikz]{bclogo}
\usepackage[english]{algorithm2e}
\usetikzlibrary{trees}
\usepackage{fancyhdr}
\usepackage{enumitem}
\usepackage[linkcolor=black, filecolor=green, colorlinks=true, urlcolor=darkgray]{hyperref}
\usepackage[left=2cm,right=2cm,top=2cm,bottom=2cm]{geometry}
\usepackage{titlesec}
\usepackage{chngcntr}
\usepackage[labelfont=bf]{caption}
\usepackage[nottoc]{tocbibind}
\usepackage{url}

\usepackage{indentfirst}

\titleformat{\section}[hang]{\bf}{\Large \thesection ~ - ~ }{0pc}{\Large}
\titleformat{\subsection}[hang]{\bf}{\large \thesubsection ~}{0pc}{\large}

\renewcommand{\sectionmark}[1]{\markboth{\uppercase{#1}}{}}

\title{\Huge \textcolor{gray}{\textit{Self-Supervised learning for Neural Architecture Search (NAS)}}}
\date{}
\author{\Large \bf Samuel Ducros}

\newcommand{\ite}{\item[$\blacktriangleright$]}

\fancyheadoffset{0cm}

\makeatletter
\let\ps@plain=\ps@fancy
\makeatother

\begin{document}

\begin{titlepage}
    \begin{sffamily}
        \begin{center}



            \begin{figure}
                \begin{minipage}[c]{.4\linewidth}
                    \includegraphics[scale=0.09]{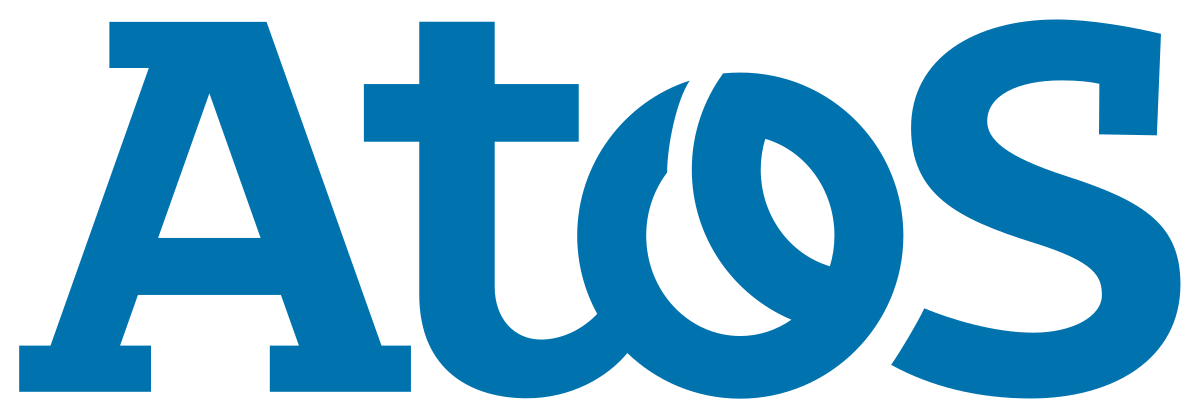}
                \end{minipage}
                \hfill
                \begin{minipage}[c]{.3\linewidth}
                    \includegraphics[scale=0.3]{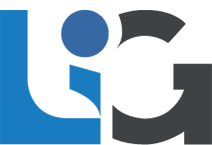}
                \end{minipage}
                \hfill
                \begin{minipage}[c]{.2\linewidth}
                    \includegraphics[scale=1.3]{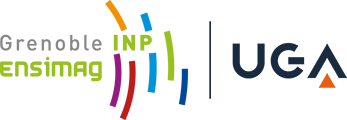}
                \end{minipage}
            \end{figure}

            \vspace{4cm}
    
            \textsc{\Large Master Thesis Report}\\[1.5cm]
            
            {\large Research internship with the LIG (Laboratoire d'Informatique de Grenoble), in partnership with Atos}\\[1.5cm]
        
            { \Huge \bfseries \textit{Self-Supervised learning for Neural Architecture Search (NAS)}\\[0.4cm] }
            
            \vspace{2cm}
        

            { \Large Samuel \textsc{Ducros}\\ M2 SIAM - DS  \\[0.4cm] }
            
            \vspace{2cm}

            \begin{minipage}[t]{0.4\textwidth}
              \begin{flushleft} \large
                \textbf{Atos}\\
                1 rue de Provence\\
                38432 Echirolles\\
                \vspace{1cm}
                \textbf{LIG}\\
                Bâtiment IMAG, 700 Avenue Centrale\\
                38401 Saint-Martin-d'Hères
              \end{flushleft}
            \end{minipage}
            \begin{minipage}[t]{0.55\textwidth}
              \begin{flushright} \large
                \emph{Tutor:} M. Loic \textsc{Pauletto}\\
                \emph{Supervisor: } M. Massih-Reza \textsc{Amini}
              \end{flushright}
            \end{minipage}

    \vfill

    {\large 15 February 2022 — 13 August 2022}

  \end{center}
  \end{sffamily}
\end{titlepage}





\tableofcontents

\newpage
\section{Introduction}

The topic of this internship is related to Self-Supervised Learning, with the main idea of finding innovative methods to train a neural network in order to make a step forward in this field. A major problem that constrains our research is the use of the smallest possible amount of annotated data to obtain good final results. The aim is to enable new AIs to understand their environment and task more efficiently and with the least amount of data possible, so that they become accessible to companies that do not have the billions of data available to Google for example. 


The objective of this internship is to propose an innovative method that uses unlabelled data, i.e. data that will allow the AI to automatically learn to predict the correct outcome. To reach this stage, the steps to be followed can be defined as follows: (1) consult the state of the art and position ourself against it, (2) come up with ideas for development paths, (3) implement these ideas, (4) and finally test them to position ourself against the state of the art, and then start the sequence again. During my internship, this sequence was done several times and therefore gives the tracks explored during the internship. 

The first sequence allowed me to get into the swing of things and to understand the subject with a large section on the state of the art. The idea that came out of it was first to speed up the execution of the code, to allow us to do tests more quickly, and at the same time to familiarise myself with the code. 


After that, and apart from the (many) incompatible computer/connection issues, I wanted to better understand how the different hyperparameters played on the results rather than doing blind tests. And there are many, which allowed me to learn a lot about the role of learning rate, norms, connections between neurons, or losses.

\subsection{Subject presentation}

The internship focuses mainly on object segmentation on an image, i.e. distinguish the different shapes and groups of shapes on an image (see \nameref{SemSeg}). Important constraints are imposed on us: 

\begin{itemize}
    \ite At the data level, when we have a dataset of images, it is very expensive to label all the images in order to train the neural network in a supervised way. We therefore want to take advantage of this dataset by keeping the major part unannotated.
    
    \ite Still at the data level, it is difficult to find a dataset of real images that corresponds to our problem. However, it is easy to obtain a large number of synthetic images, extracted from a recent and realistic video game for example.
    
    \ite At the hardware level, we have access to several GPUs and CPUs that will allow us to accelerate our training.
\end{itemize}



Considering these constraints and our goal, we base our work on the one hand on the research of innovative methods of Semi-Supervised Learning to take into account unlabeled data, and on the other hand on the Domain Adaptation, that is to say the fact of training our network for the segmentation of images on synthetic images, of which we have a lot of data. 

This report is based on the first part of the problematic on the research in Semi-Supervised Learning for the consideration of unlabeled data. A major part concerns the knowledge of the state of the art around Semi-Supervised Learning techniques and around Semantic Segmentation. 




\section{Subject development}

\subsection{Technologies}

The technologies presented below were used for all the work I was able to do. When I joined the team, the project has already been developed with the Python language and various libraries, such as Pytorch and OpenCV. So these are the technologies I continued with.

\subsubsection{Python}

Python\footnote{\url{https://www.python.org}} is the most popular language in the world of data analysis (data science) and artificial intelligence (machine learning, deep learning). This craze is reflected in the large number of libraries that allow the manipulation of mathematical objects or concepts with a high level of positioning.
Among these, we can note the use of NumPy which allows manipulating matrices, or Scikit-learn which proposes machine learning bricks and matplotlib for all the aspects concerning the visualization.
 
As the language is simple to use, untyped, and does not have a long compilation time, we can quickly iterate by changing different parameters and approaches. This is ideal, as the goal is not to have an optimized version of the work since it is in the research state.

\subsubsection{Pytorch}

PyTorch\footnote{\url{https://pytorch.org}} is an open source Python machine learning software library based on Torch developed by Facebook. PyTorch allows the tensor calculations necessary for deep learning to be performed. These calculations are optimized and carried out either by the processor (CPU) or, where possible, by a graphics processor (GPU) supporting CUDA. It comes from the research teams at Facebook, and before that from Ronan Collobert in Samy Bengio's team at IDIAP. PyTorch is derived from an earlier software, Torch, which was used with the Lua language. PyTorch is independent of Lua and is programmed in Python.

\subsubsection{OpenCV}

OpenCV\footnote{\url{https://opencv.org}} is a library for image and video processing. It is a library written in C++ that also supports CUDA calls and offers a Python API to facilitate development.
It includes many of the algorithms we use, such as object tracking within a video, image modification when preprocessing our data, or simply capturing or displaying video content to highlight our system's detections.

\newpage

\subsection{State of the art}

A great part of my internship was to understand the relationships between all the possible components of a neural network, with all the parameters and hyperparameters. Each advance in understanding brought up a new questioning on another point. So a lot of research has been done. First of all, I had to understand the subject and research the state of the art in general on Deep Learning and Semantic Segmentation, which are the very first basis of the subject. More precisely, what we are interested in is to use as little annotated data as possible, and therefore we are interested in Self-Supervised Learning and Semi-Supervised Learning. Here are my research results concerning Semantic Segmentation, Self and Semi-Supervised Learning.


    \subsubsection{Semantic Segmentation}
    
\label{SemSeg}

\begin{figure}[h!]
    \centering
    \includegraphics[width=0.6\linewidth]{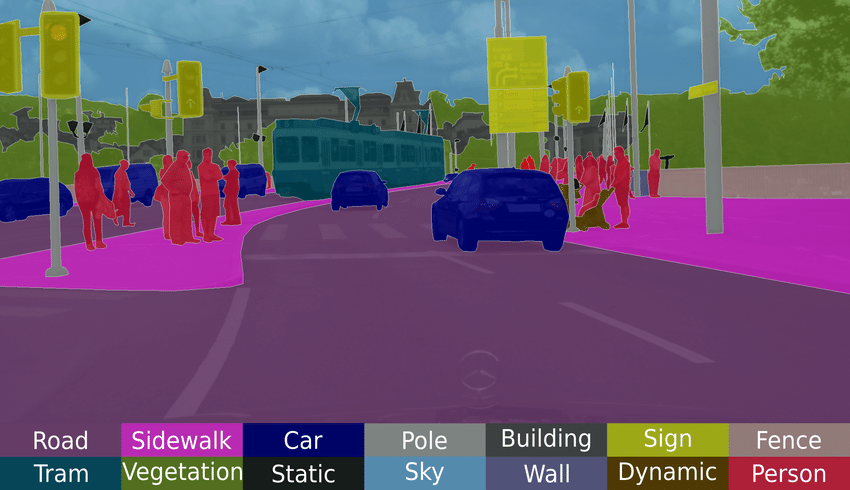}
    \caption{Semantic segmentation of a scene from the Cityscapes dataset by Cordts et al. (2016) recorded in Zurich.}
    \label{fig:SemSegCity}
\end{figure}

Semantic segmentation, or image segmentation, is the task of clustering parts of an image together which belong to the same object class. It is a form of pixel-level prediction because each pixel in an image is classified according to a category. Some example benchmarks for this task are Cityscapes\footnote{\url{https://www.cityscapes-dataset.com}}  (see \textbf{Figure \ref{fig:SemSegCity}}), PASCAL VOC\footnote{\url{http://host.robots.ox.ac.uk/pascal/VOC/}} and ADE20K\footnote{\url{https://groups.csail.mit.edu/vision/datasets/ADE20K/}}. Models are usually evaluated with the Mean Intersection-Over-Union (mIoU, see \nameref{mIoU}).

\begin{figure}[h!]
    \centering
    \includegraphics[width=\linewidth]{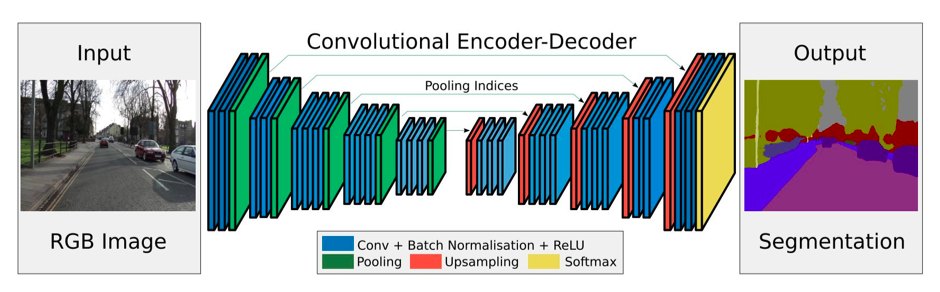}
    \caption{Example of architecture for image segmentation of road scenes \cite{badrinarayanan2016segnet}}
    \label{fig:SegEncDec}
\end{figure}

The basic architecture in image segmentation consists of an encoder and a decoder (see \textbf{Figure \ref{fig:SegEncDec}}). The encoder extracts features from the image through filters. The decoder is responsible for generating the final output which is usually a segmentation mask containing the outline of the object. Most of the architectures have this architecture or a variant of it.

    \subsubsection{Semi-Supervised Learning}


Supervised learning usually requires a large amount of labelled data. Obtaining good quality labelled data is a costly and time-consuming task, especially for a complex task such as object detection, instance segmentation and more detailed annotations are desired. On the other hand, unlabelled data is readily available in abundance. 

Semi-supervised learning (SSL), also known as learning with partially labelled data, refers to the process of learning a prediction function from labelled and unlabelled training samples.  In this situation, the labelled instances are expected to be few in number, resulting in an inefficient supervised model, but the unlabelled training examples contain useful information about the prediction problem at hand, which can be exploited to produce an efficient prediction function. We assume that a collection of labelled training examples derived from a joint probability distribution and a collection of unlabelled training examples derived from the marginal distribution are both accessible in this case. The problem arises in supervised learning if the unlabelled data set is empty. The opposite extreme example is when the labelled training set is empty, in which case the problem is reduced to unsupervised learning.\\

\textbf{Smoothness} is a fundamental assumption of semi-supervised learning, which states that two instances in a high density region must have identical class labels. This means that if two points are part of the same group or cluster, their class labels will most likely be the same. On the other hand, if they are separated by a low density area, their desired labels should be different.

Suppose that the instances of the same class form a partition; the unlabelled training data could help to determine the partition boundary more efficiently than if only labelled training examples were used. Therefore, searching for partitions using a mixture model and then assigning class labels to groups using the labelled data they comprise is a technique for using unlabelled data to train a model.
If two instances are in the same group, it is likely that they belong to the same class, according to the underlying assumption, known as the \textbf{cluster assumption}. This assumption can be explained as follows: if a group is created by a large number of instances, it is rare that they all belong to the same class. This does not imply that a class consists of only one group of instances, but rather that two instances of distinct classes are unlikely to be in the same group. 
If we consider the example partitions as high density regions, another version of the cluster assumption is that the decision boundary passes through low density regions, according to the previous smoothing assumption.

Density estimation is often based on a notion of distance, which may not make sense in high-dimensional vector spaces. To resolve this difficulty, a third assumption known as the \textbf{manifold assumption}, which is supported by a number of semi-supervised models, holds that instances in high-dimensional spaces exist on low-dimensional topological spaces that are locally Euclidean (or geometric manifolds).\\

Self-training is one of the first wraparound techniques for learning a supervised classifier using partially labelled data. A supervised method is first trained on the labelled training set, and its predictions are then used to assign pseudo-labels to a portion of the unlabelled training samples. The supervised classifier is then retrained on the augmented training set (labelled and pseudo-labelled), and the procedure is repeated until there are no unlabelled observations left to pseudo-label. Despite its simplicity, self-training is difficult to analyse in general. Some studies have proposed bounds on the error of majority-vote classifiers, used in the envelope, on unlabelled training data. When the majority-voting classifier makes most of its errors on low-density regions, this bound is shown to be tight \cite{FeofanovDA19,NIPS2008_dc6a7071,UsunierAG11,Krithara:08}.

\begin{figure}[h!]
    \centering
    \includegraphics[width=\linewidth]{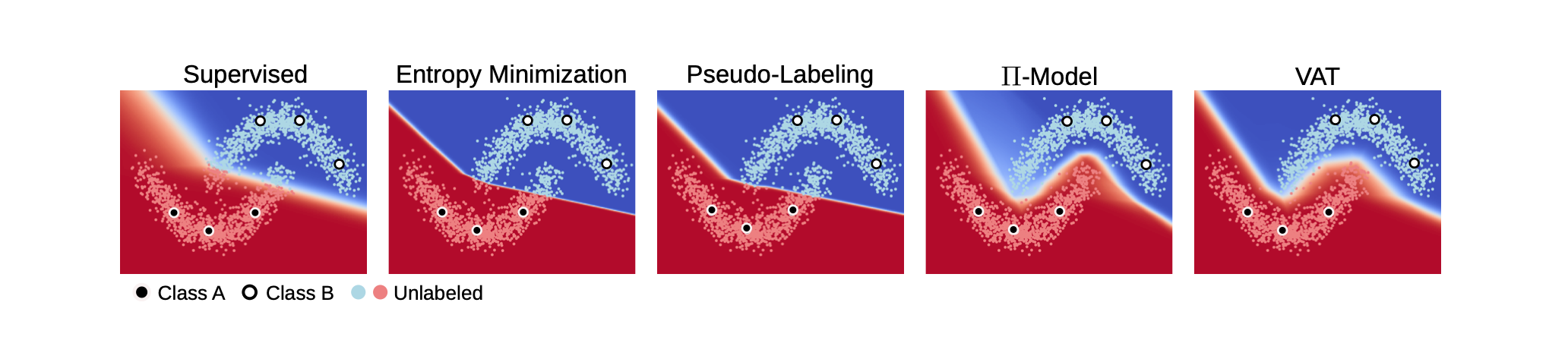}
    \caption{SSL toy example. The decision boundaries obtained on two-moons dataset, with a supervised and different SSL approaches using 6 labeled examples, 3 for each class, and the rest of the points as unlabeled data. (Source: \cite{ouali2020overview})}
    \label{fig:SSL}
\end{figure}

The unsupervised learning method we used to take into account the unlabelled data is called the Self-supervised Learning \cite{Weng_2019}. It consists in creating the input data and the target data from the unlabelled data to provide the supervision. This task could be as simple as given the upper-half of the image, predict the lower-half of the same image \cite{yu2018generative}, or given the grayscale version of the colored image, predict the LAB channels (for the CIELAB Color Space \cite{cielab}) of the same image \cite{zhang2016colorful}.

In \cite{zhai2019s4l}, the authors show the advantage of using self-supervised learning in the context of semi-supervised learning, by introducing the Self-Supervised Semi-Supervised Learning (S4L) framework to derive two new methods for semi-supervised image classification.\\

Lately, in natural language processing, Transformer models \cite{Weng_2020} have achieved a lot of success. Transformers like Bert \cite{devlin2019bert} or T5 \cite{raffel2020exploring} applied the idea of self-supervision to NLP (Natural Language Processing) tasks. They first train the model with large unlabelled data and then fine-tuning the model with few labelled data examples.\\



\begin{figure}[h!]
    \centering
    \includegraphics[width=\linewidth]{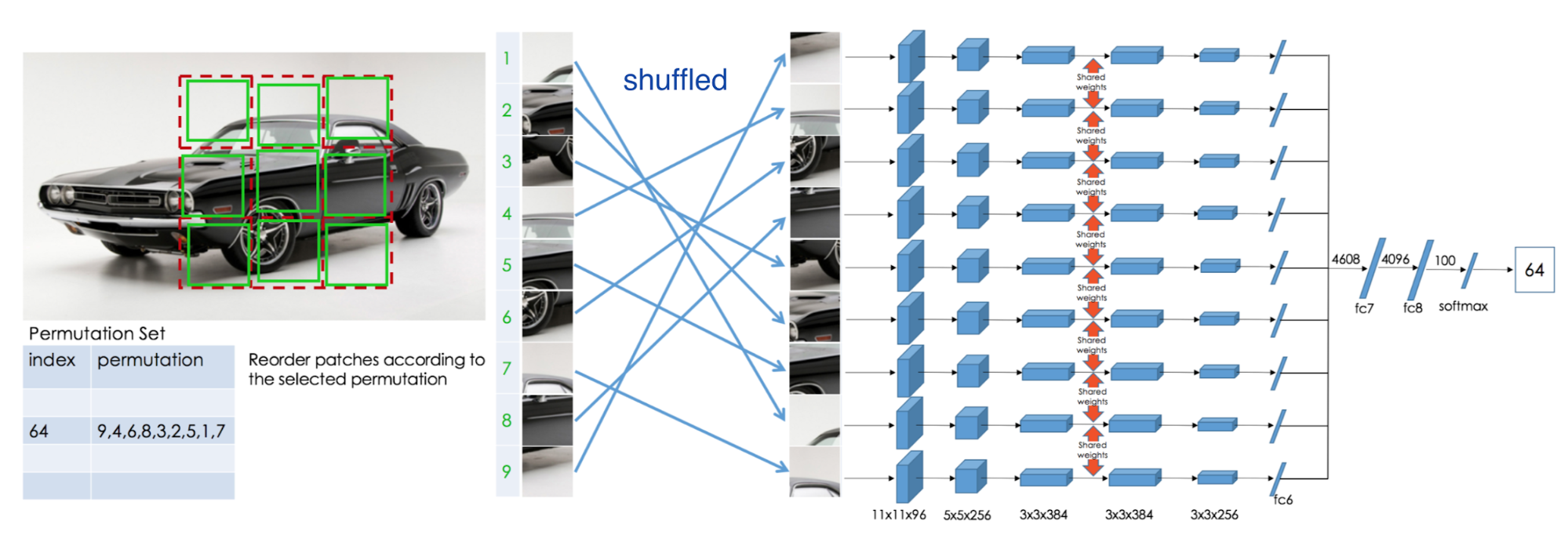}
    \caption{Illustration of self-supervised learning by solving jigsaw puzzle (Source: \cite{noroozi2017unsupervised})}
    \label{fig:SelfSupJig}
\end{figure}



On this basis, I further refine my research by orienting it towards Multi-Task Learning and Transfer Learning. The topic leads us to think about the use of unannotated data to allow the learning of the Segmentation task. The idea is indeed to train our network with unsupervised pretext-tasks in order to transfer knowledge for the learning of our target task. 


    \subsubsection{Transfer Learning}
    
Transfer learning \cite{zhuang2020comprehensive} is one of the research fields in machine learning that aims to transfer knowledge from one or more source tasks to one or more target tasks. It can be seen as the ability of a system to recognize and apply knowledge and skills, learned from previous tasks, to new tasks or domains sharing similarities.

\begin{figure}[h!]
    \centering
    \includegraphics[width=0.7\linewidth]{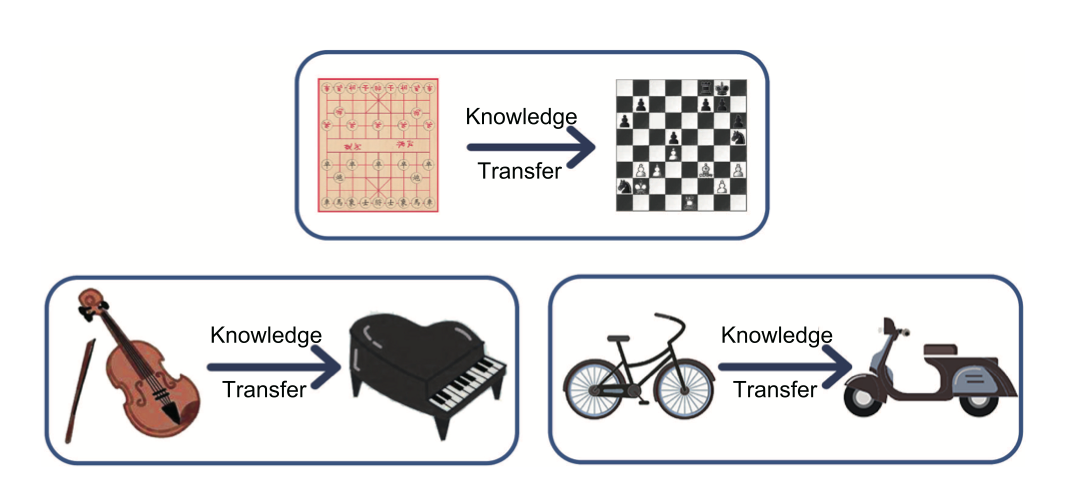}
    \caption{Intuitive examples about transfer learning. (Source: \cite{zhuang2020comprehensive})}
    \label{fig:Transfer}
\end{figure}

    \subsubsection{Multi-Task Learning}
    
Multi-task learning (MTL) \cite{crawshaw2020multitask} is a subfield of machine learning in which multiple learning tasks are solved at the same time, while exploiting commonalities and differences across tasks. This can result in improved learning efficiency and prediction accuracy for the task-specific models, when compared to training the models separately. Early versions of MTL were called "hints".

In a widely cited 1997 paper, Rich Caruana gave the following characterization:\\

\textit{"Multitask Learning is an approach to inductive transfer that improves generalization by using the domain information contained in the training signals of related tasks as an inductive bias. It does this by learning tasks in parallel while using a shared representation; what is learned for each task can help other tasks be learned better."}\\


\begin{figure}[h!]
    \centering
    \includegraphics[width=\linewidth]{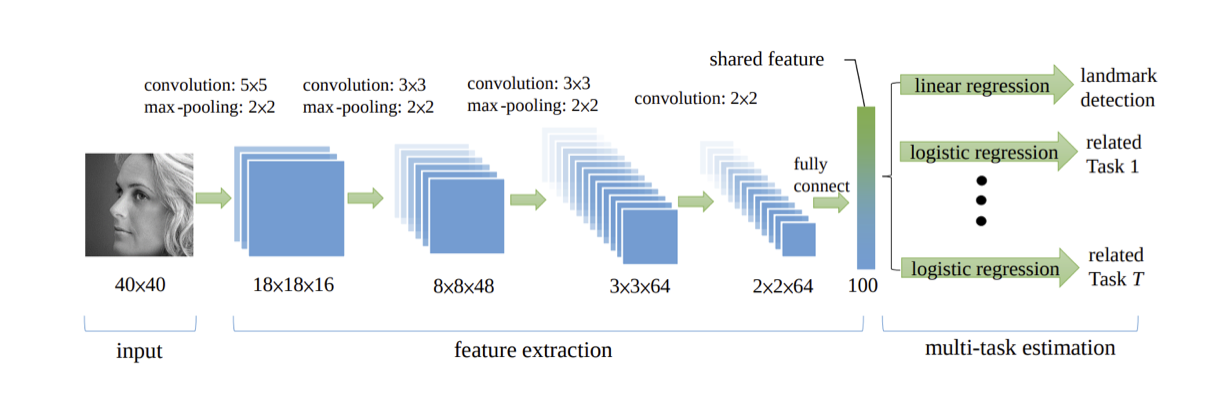}
    \caption{Architecture for TCDCN \cite{chen2020deep}. The base feature extractor is made of a series of convolutional layers which are shared between all tasks, and the extracted features are used as input to task-specific output heads. (Source: \cite{crawshaw2020multitask})}
    \label{fig:MultiTasks}
\end{figure}

Multi-task learning works because regularization induced by requiring an algorithm to perform well on a related task can be superior to regularization that prevents overfitting by penalizing all complexity uniformly. One situation where MTL may be particularly helpful is if the tasks share significant commonalities and are generally slightly under sampled.\\

\subsection{Comprehension and getting started with the code}
    
The first code I did on this project was mainly optimization. Indeed, as it was, the code allowed to launch experiments using only one pretext-task among Inpainting and Denoising, with 3 supervised and 3 unsupervised images. By understanding more in depth each part of the code, I could find pieces of code repeated at each iteration unnecessarily, or (and especially) data stored unnecessarily or which accumulated. From a program that took about 10 days to run with a configuration C that accepted at most 3 supervised and 3 unsupervised images, we went to a program that runs for 3 days that accepts up to 8 supervised and 8 unsupervised images for the same configuration C. At the same time, I rewrote some of the code to make it more automatic and general, and to anticipate future improvements, for example with for loops on pretext-tasks instead of the if statement, to generalise all pretext-task possibilities.\\

Now taking these optimizations as a base, I improved the code so that it can take into account several pretext-tasks at the same time for the training. Thus the input images undergo different transformations depending on the pretext-task. 

        \subsubsection{Inpainting}

Existing works for image inpainting \cite{yu2018generative} can be mainly divided into two groups. The first group represents traditional diffusion-based or patch-based methods with low-level features. The second group, that we use, attempts to solve the inpainting problem by a learning-based approach, e.g. training deep convolutional neural networks to predict pixels for the missing regions.

As described above, the network that learns the inpainting task is constituted of an common encoder with the other tasks, and a decoder head of its own. The input image is the grand truth image from which we erase a square. After having got through the encoder and the decoder, the prediction is compared with the grand truth image using Mean Squared Error Loss (see \nameref{mseloss}).

\begin{figure}[h!]
    \centering
    \includegraphics[width=\linewidth]{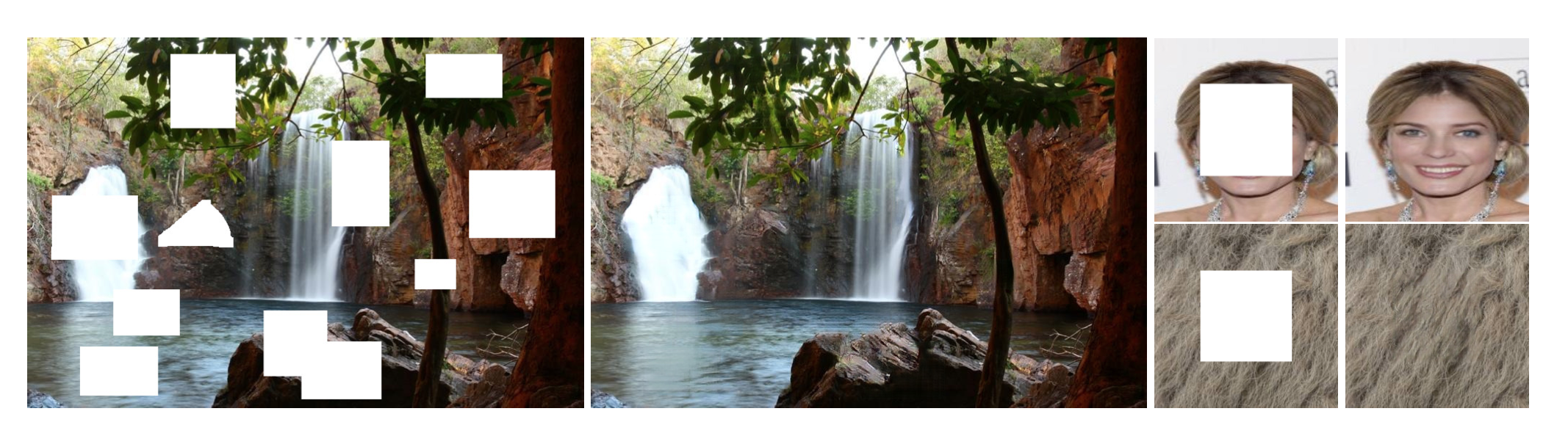}
    \caption{Example inpainting results of the method of \cite{yu2018generative} on images of natural scene, face and texture. Missing regions are shown in white. In each pair, the left is input image and right is the direct output of the trained generative neural networks without any post-processing. (Source: \cite{yu2018generative})}
    \label{fig:Inp}
\end{figure}
    
        \subsubsection{Noise}
        
Many different ways of denoising methods exist \cite{tian2020deep}. From filters to deep learning, many researches have been done in this area. Here we use deep learning and, as the inpainting task, we build an encoder network and a decoder network specific for the denoising task. The input image is the grand truth which we added a virtual noise. After having got through the network, the prediction is compared to this grand truth using Mean Squared Error Loss (see \nameref{mseloss}).

\begin{figure}[h!]
    \centering
    \includegraphics[width=\linewidth]{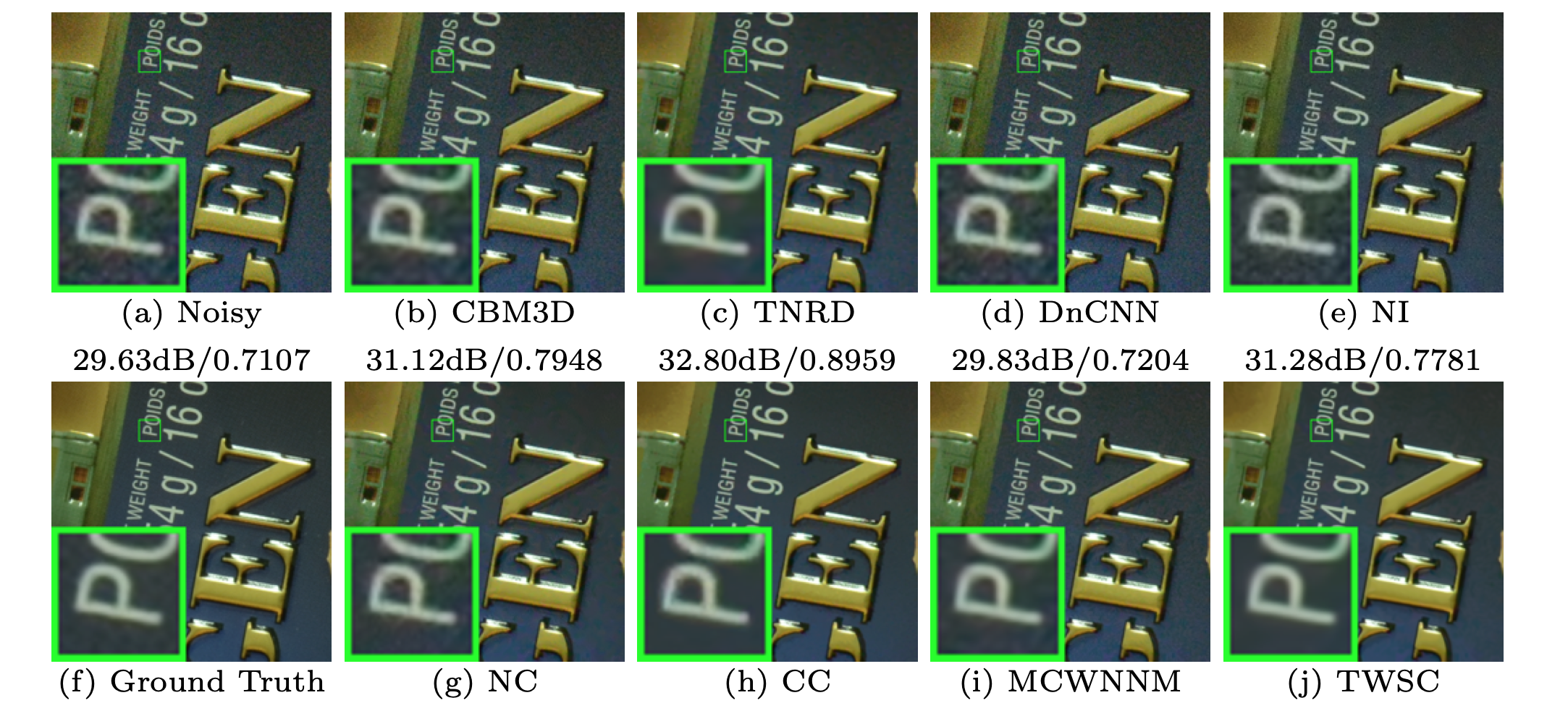}
    \caption{Denoised images of the real noisy image Nikon D800 ISO 6400 1 \cite{7780555} by different methods (Source: \cite{xu2018trilateral})}
    \label{fig:Denoise}
\end{figure}





    
    
        \subsubsection{Colorization} 

Colorization \cite{zhang2016colorful} is the process of adding plausible color information to monochrome photographs or videos. As the denoising problem, there are several ways to resolve this problem \cite{vitoria2020chromagan}. Here we need to train the colorization with our network. As before, a decoder head is specific to the colorization task. To learn it in a unsupervised way, we give as input a colored image which we transformed in a grayscale image. The prediction colors are compared with the grand truth with  
    
        \subsubsection{Jigsaw} 

\begin{figure}[h!]
    \centering
    \includegraphics[width=\linewidth]{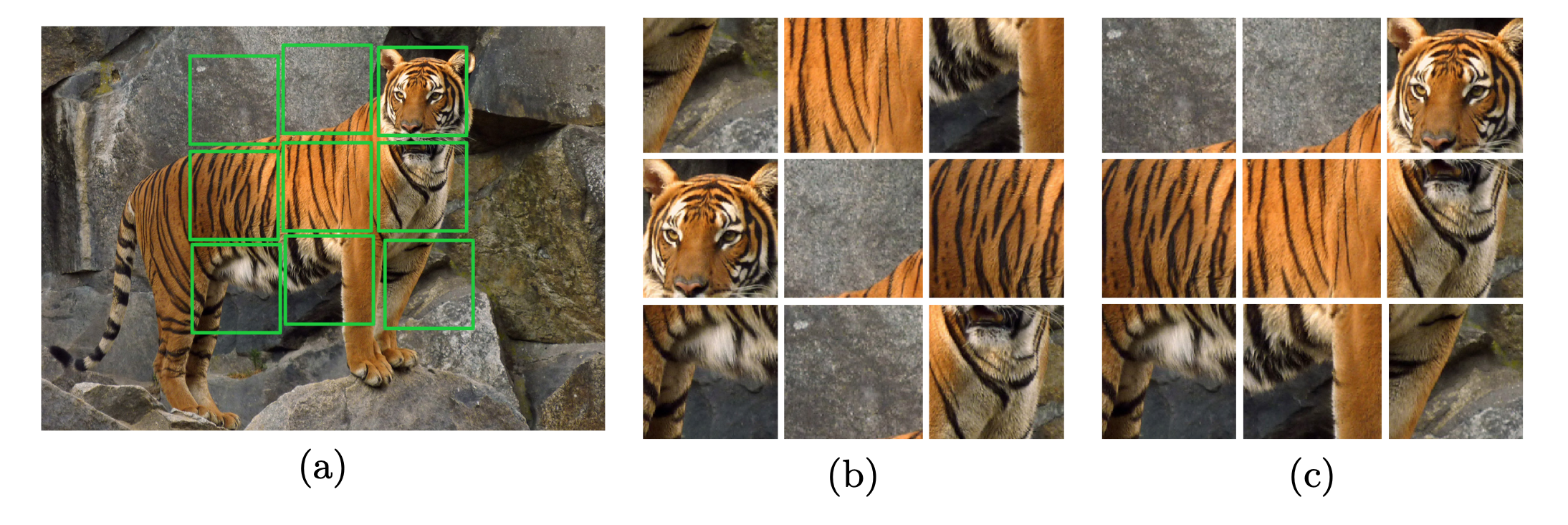}
    \caption{Learning image representations by solving Jigsaw puzzles. (a) The image from which the tiles (marked with green lines) are extracted. (b) A puzzle obtained by shuffling the tiles. Some tiles might be directly identifiable as object parts, but others are ambiguous (e.g., have similar patterns) and their identification is much more reliable when all tiles are jointly evaluated. In contrast, with reference to (c), determining the relative position between the central tile and the top two tiles from the left can be very challenging. (Source: \cite{noroozi2017unsupervised})}
    \label{fig:Jigsaw}
\end{figure}

The Jigsaw problem \cite{noroozi2017unsupervised} is the process of solving a puzzle of an image, i.e. find the original order between the different tiles extracted from an image (see \textbf{Figure \ref{fig:Jigsaw}}). Its decoder head is of the same type as the decoder head of a classification problem, because the problem here is to well classify every part of the puzzle.

\subsection{Tuning the Hyper parameters}
    
Now that we have these new features and the code works well and is optimized, we want to run tests and get the best possible results. But before that, we have to choose the right hyperparameters to test. It took a lot of research to understand their impact and to know the different possible arrangements. Here I summarize my research results on the main hyperparameters.
    
    \subsubsection{Normalization techniques}

Normalization techniques can decrease your model’s training time by a huge factor. It normalizes each feature so that they maintains the contribution of every feature, as some feature has higher numerical value than others. This way our network can be unbiased (to higher value features). Batch Norm for example makes loss surface smoother (i.e. it bounds the magnitude of the gradients much more tightly) \cite{santurkar2019does}. It makes the Optimization faster because normalization does not allow weights to explode all over the place and restricts them to a certain range. A last unintended benefit of Normalization is that it helps network in Regularization (only slightly, not significantly). Therefore getting Normalization right can be a crucial factor in getting your model to train effectively.

Let’s dive into details of each normalization technique one by one.\\
\begin{itemize}

    \item \textbf{Batch Normalization:} Batch normalization \cite{Aakash_2019} is a method that normalizes activations in a network across the mini-batch of definite size. For each feature, batch normalization computes the mean and variance of that feature in the mini-batch. It then subtracts the mean and divides the feature by its mini-batch standard deviation. We can add $\gamma$ and $\beta$ as scale and shift learn-able parameters respectively, in order to take into account a greater magnitude of the weights if necessary. This all can be summarized as:
    
    Let $\mathcal{B} = \{ x_{1 ... m} \}$ be the mini-batch, containing the features of each data of the batch. Let $\gamma$ and $\beta$ two parameters to learn. We have, with $\epsilon$ is the stability constant in the equation:
    
    \begin{align*}
        \mu_{\mathcal{B}} &= \dfrac{1}{m} \sum_{i=1}^m x_i\\
        \sigma_{\mathcal{B}}^2 &= \dfrac{1}{m} \sum_{i=1}^m (x_i - \mu_{\mathcal{B}})^2\\
        \hat x_i &= \dfrac{x_i - \mu_{\mathcal{B}}}{\sqrt{\sigma_{\mathcal{B}}^2 + \epsilon}}\\
        BN_{\gamma, \beta}(x_i) &= \gamma \hat x_i + \beta
    \end{align*}

    \item \textbf{Layer Normalization:} Layer Normalization differs from the Batch Normalization because it normalizes input across the features instead of normalizing input features across the batch dimension.
    
    \item \textbf{Instance(or Contrast) Normalization:} Layer normalization and Instance normalization are very similar to each other but the difference between them is that Instance normalization normalizes across each channel in each training example instead of normalizing across input features in an training example.

    \item \textbf{Group Normalization:} Group Normalization normalizes over group of channels for each training example.
\end{itemize}

    

\begin{figure}[h!]
    \centering
    \includegraphics[width=\linewidth]{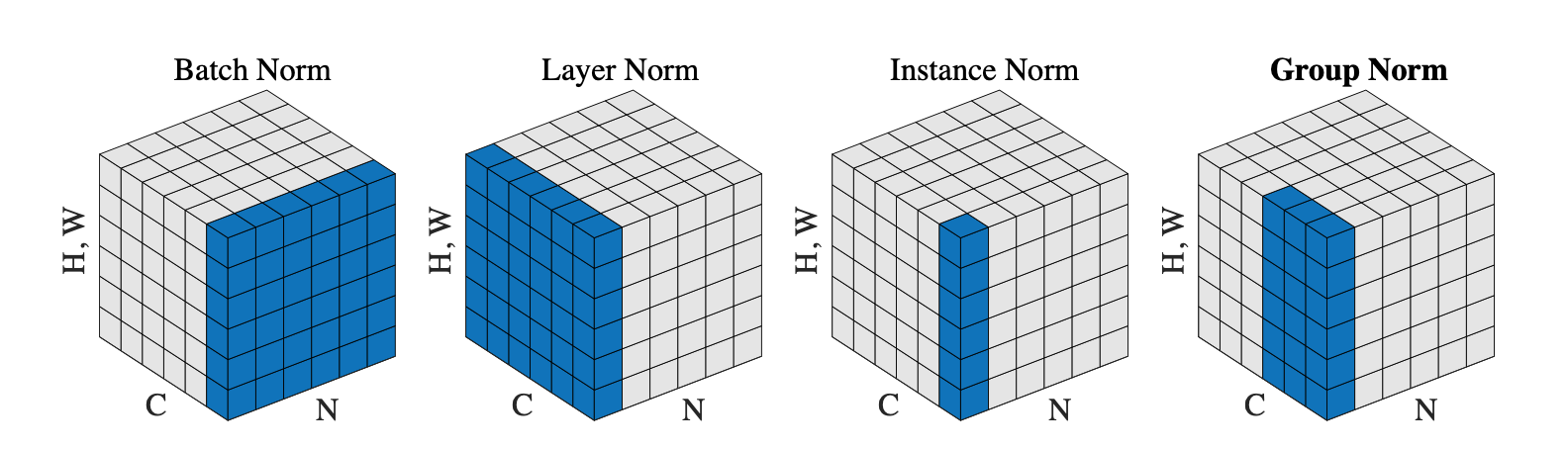}
    \caption{Normalization methods. Each subplot shows a feature map tensor, with N as the batch axis, C as the channel axis, and (H, W ) as the spatial axes. The pixels in blue are normalized by the same mean and variance, computed by aggregating the values of these pixels. (Source: \cite{wu2018group})}
    \label{fig:Norms}
\end{figure}

We summarize here these norms in the schema in \textbf{Figure \ref{fig:Norms}} \cite{wu2018group}. For the project we mainly employed \textbf{Switchable Normalization (SN)}\cite{luo2018normalization}, which is a normalization method that uses a weighted average of different mean and variance statistics from batch normalization, instance normalization, and layer normalization. Switch Normalization can outperform Batch normalization on tasks such as image classification and object detection. \cite{luo2018normalization} shows that the instance normalization is used more often in earlier layers, batch normalization is preferred in the middle and layer normalization being used in the last more often. Smaller batch sizes lead to a preference towards layer normalization and instance normalization.

    \subsubsection{Losses}
    
In the context of an optimization algorithm, the function used to evaluate a candidate solution (i.e. a set of weights) is referred to as the objective function. We may seek to maximize or minimize the objective function, meaning that we are searching for a candidate solution that has the highest or lowest score respectively. Typically, with neural networks, we seek to minimize the error. As such, the objective function is often referred to as a cost function or a loss function and the value calculated by the loss function is referred to as simply “loss.”

The cost function reduces all the various good and bad aspects of a possibly complex system down to a single number, a scalar value, which allows candidate solutions to be ranked and compared. In calculating the error of the model during the optimization process, a loss function must be chosen. This can be a challenging problem as the function must capture the properties of the problem and be motivated by concerns that are important to the project and stakeholders. It is important, therefore, that the function faithfully represent our design goals. If we choose a poor error function and obtain unsatisfactory results, the fault is ours for badly specifying the goal of the search.

Here are the different loss functions we use for our project, depending on the task that is trained.

\begin{itemize}
    \ite \textbf{Cross-Entropy:} When modeling a classification problem where we are interested in mapping input variables to a class label, we can model the problem as predicting the probability of an example belonging to each class. In a binary classification problem, there would be two classes, so we may predict the probability of the example belonging to the first class. In the case of multiple-class classification, we can predict a probability for the example belonging to each of the classes. In the training dataset, the probability of an example belonging to a given class would be 1 or 0, as each sample in the training dataset is a known example from the domain. We know the answer. Therefore, under maximum likelihood estimation, we would seek a set of model weights that minimize the difference between the model’s predicted probability distribution given the dataset and the distribution of probabilities in the training dataset. This is called the cross-entropy. For the project, the cross-entropy is used as loss function for tasks as Jigsaw, Colorization and Semantic Segmentation. 
    
    \ite \label{mseloss} \textbf{Mean Squared Error (MSE):} Mean squared error (MSE) is the most commonly used loss function for regression. The loss is the mean overseen data of the squared differences between true and predicted values. MSE is sensitive towards outliers and given several examples with the same input feature values, the optimal prediction will be their mean target value. MSE is thus good to use if you believe that your target data, conditioned on the input, is normally distributed around a mean value, and when it’s important to penalize outliers extra much. We use MSE when doing regression, believing that your target, conditioned on the input, is normally distributed, and want large errors to be significantly (quadratically) more penalized than small ones. For the project then, MSE is used as loss function for tasks as Inpainting and Denoising.
\end{itemize}

    \subsubsection{Learning rate / Optimizer} 
    
\begin{figure}[h!]
    \centering
    \includegraphics[width=\linewidth]{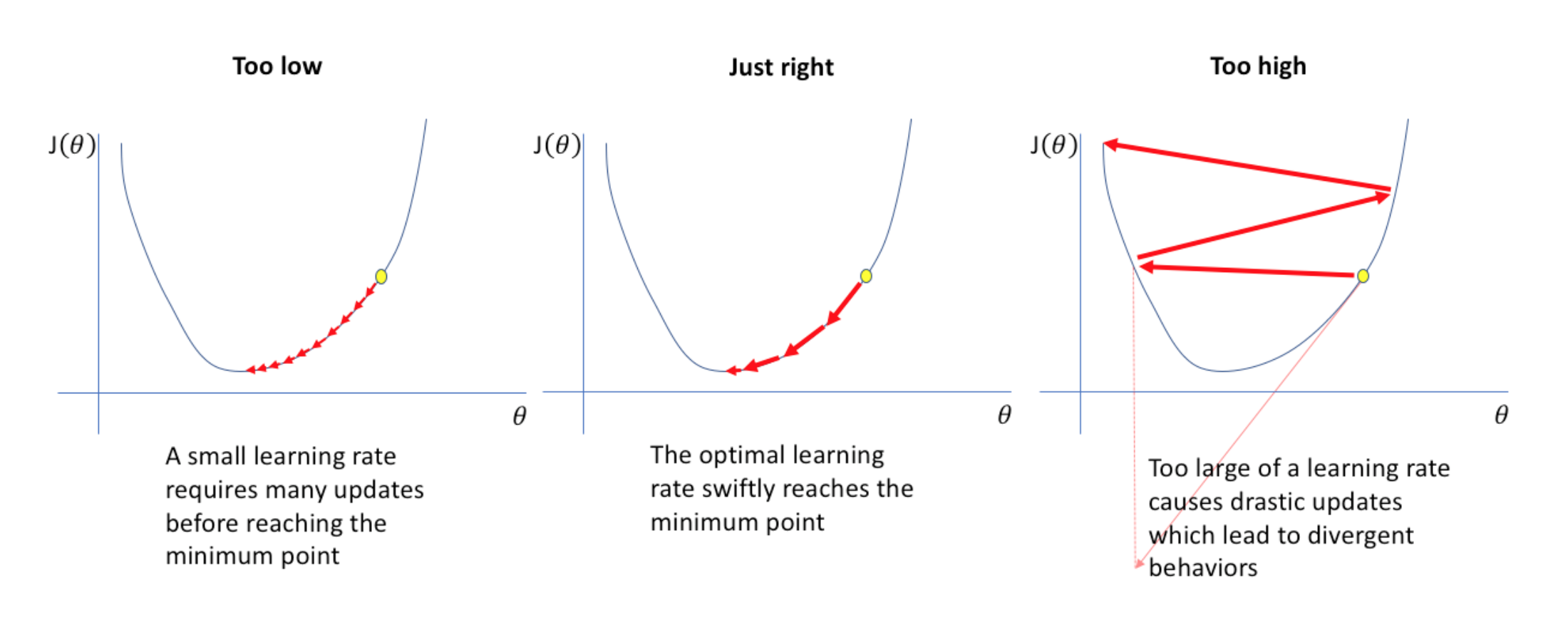}
    \caption{Consequences of the learning rate (Source: \cite{Jordan_2018})}
    \label{fig:LR}
\end{figure}

The learning rate \cite{Jordan_2018} is a hyperparameter that controls how much to change the model in response to the estimated error each time the model weights are updated. Choosing the learning rate is challenging as a value too small may result in a long training process that could get stuck, whereas a value too large may result in learning a sub-optimal set of weights too fast or an unstable training process (see \textbf{Figure \ref{fig:LR}}).

The stochastic gradient descent (SGD) algorithm is a common optimizer algorithm for the training. SGD is an optimization algorithm that estimates the error gradient for the current state of the model using examples from the training dataset, then updates the weights of the model using the back-propagation of errors algorithm, referred to as simply backpropagation. Other learning rate optimization algorithms can be used as ADAM \cite{Bushaev_2018}.

The learning rate controls how quickly the model is adapted to the problem. Smaller learning rates require more training epochs given the smaller changes made to the weights each update, whereas larger learning rates result in rapid changes and require fewer training epochs. A learning rate that is too large can cause the model to converge too quickly to a suboptimal solution, whereas a learning rate that is too small can cause the process to get stuck. The challenge of training deep learning neural networks involves carefully selecting the learning rate. It may be the most important hyperparameter for the model.

To best chose the learning, we use a learning rate schedule, that changes the learning rate during learning and is most often changed between epochs/iterations. This is mainly done with two parameters: decay and momentum. Decay serves to settle the learning in a nice place and avoid oscillations, a situation that may arise when a too high constant learning rate makes the learning jump back and forth over a minimum, and is controlled by a hyperparameter. Momentum is analogous to a ball rolling down a hill; we want the ball to settle at the lowest point of the hill (corresponding to the lowest error). Momentum both speeds up the learning (increasing the learning rate) when the error cost gradient is heading in the same direction for a long time and also avoids local minima by 'rolling over' small bumps. Momentum is controlled by a hyper parameter analogous to a ball's mass which must be chosen manually—too high and the ball will roll over minima which we wish to find, too low and it will not fulfil its purpose.\\

For our project we use mainly the SGD optimization algorithm.

\section{Train Protocol}

Finally we have a functional code and we understand the role of hyperparameters. Now it's time to launch tests. Our goal is to compare ourselves to the state of the art regarding the segmentation results, and to make our method exceed these results. Let's explain how we compare.

    \label{mIoU}
    \subsection{mIoU (mean Intersection over Union)}

There are several neural network models working on different platforms, and different unique approaches for object detection and semantic image segmentation, so we need to know how to choose one among all in order to have better results in our field. There has to be a criterion based on which such decision can be made. The best one is by checking the degree of similarity of the output produced by such methods with the ground truth and that can be done in a mathematical way by calculating IoU (Intersection over Union) between the two. This method takes into account the region common to both (ground truth and predicted output) and computes to what percentage it has similarity with the actual one.

It’s quite simple in case of “Single-class based Semantic Image Segmentation” but not in the case of other “Multiple-class based Semantic Image Segmentation”, as in the case of Pascal VOC challenge (with 21 classes) where there can be objects belonging to different classes in the same image. In such cases each object has to be given a different label and has to be treated accordingly during IoU computation. A method which can take into account such cases and find out the overall IoU for multiple classes present in an image is the calculation of the mean value of IoUs corresponding to different classes which would match with the actual degree of similarity. This mean value is regarded as mean IoU (mIoU).

Here is a common approx for the computation of the mIoU, which we use in our code. For that we need the labelled matrix of both predicted result and expected one (ground truth). Let two matrices $GT$ and $Pred$, one representing the actual segmented output and the other predicted by any neural network or model. The elements of these matrices are the labels representing different classes to which pixels, at that particular location on the image belong. Then here are the steps to follow:

\begin{itemize}
    \ite Finding out the frequency count of each class for both the matrix, $F_{GT}$ and $F_{Pred}$,
    \ite Converting the matrices to 1D format, $GT_{1D}$ and $Pred_{1D}$
    \ite Finding out the category matrix, of size $(nb_{classes} \times nb_{classes})$. The category matrix is one that will have the elements as the category numbers to which the pixels at that particular location belong. 
    $Categ = (nb_{classes} \times GT_{1D}) + Pred_{1D}$
    \ite Constructing the confusion matrix. A confusion matrix is a $(nb_{classes} \times nb_{classes})$ size matrix which stores the information about the number of pixels belonging to a particular category. The frequency count of the ‘category’ array gives a linear array which on reshaping to $(nb_{classes} \times nb_{classes})$ gives us the confusion matrix. 
    $CM = Categ.reshape((nb_{classes} \times nb_{classes}))$
    \ite Calculating IoU for individual classes. The diagonal of the confusion matrix represents the common region. So, these elements are the intersection values of the predicted output and ground truth.
    $I = diag(CM)$
    $U = GT_{1D} + Pred_{1D} - I$
    \ite Calculating MIoU for the actual-predicted pair.
    $IoU = \dfrac{I}{U}$ is a matrix of size $nb_classes$. $mIoU = mean(IoU)$ is the mean over all values of $IoU$.
\end{itemize}

In case of multiple classes, the mIoU has to be calculate rather than just calculating IoU by treating all the different classes as a single one. So, considering all the classes mIoU has to be calculated for validation.\\

There are several other methods to find out the similarity between an actual image and predicted result, most popular among them is the bounding box method \cite{adhikari2020iterative} but the instances involving fine edge detection along with segmentation where higher accuracy would be required, this method proves to be reliable.

\newpage
\bibliographystyle{unsrt}
\bibliography{ref}

    
    

\end{document}